\documentclass[10pt,twocolumn,letterpaper]{article}

\usepackage{wacv}
\usepackage{times}
\usepackage{epsfig}
\usepackage{graphicx}
\usepackage{amsmath}
\usepackage{amssymb}

\usepackage{color}
\usepackage{bm}
\usepackage{multirow}
\usepackage{booktabs}
\usepackage[pagebackref=true,breaklinks=true,letterpaper=true,colorlinks,bookmarks=false]{hyperref}

\def\etal{\emph{et al.}}

\def\wacvPaperID{137} 

\wacvfinalcopy 

\ifwacvfinal
\def\assignedStartPage{1} 
\fi

\ifwacvfinal
\usepackage[breaklinks=true,bookmarks=false]{hyperref}
\else
\usepackage[pagebackref=true,breaklinks=true,colorlinks,bookmarks=false]{hyperref}
\fi

\ifwacvfinal
\else
\fi

\begin{document}

\title{Audio-Visual Event Localization via Recursive Fusion by Joint Co-Attention}

\author{Bin Duan$^1$\quad Hao Tang$^2$\quad Wei Wang$^3$\quad Ziliang Zong$^1$\quad Guowei Yang$^1$\quad Yan Yan$^1$\\
$^1$Department of Computer Science, Texas State University, USA\\
$^2$DISI, University of Trento, Italy\quad\, $^3$CVLab, EPFL, Switzerland\\
\tt\small \{bin.duan, ziliang, gyang, tom\_yan\}@txstate.edu, hao.tang@unitn.it, wei.wang@epfl.ch
}

\maketitle

\begin{abstract}
The major challenge in audio-visual event localization task lies in how to fuse information from multiple modalities effectively. Recent works have shown that attention mechanism is beneficial to the fusion process. In this paper, we propose a novel joint attention mechanism with multimodal fusion methods for audio-visual event localization. Particularly, we present a concise yet valid architecture that effectively learns representations from multiple modalities in a joint manner. Initially, visual features are combined with auditory features and then turned into joint representations. Next, we make use of the joint representations to attend to visual features and auditory features, respectively. With the help of this joint co-attention, new visual and auditory features are produced, and thus both features can enjoy the mutually improved benefits from each other. It is worth noting that the joint co-attention unit is recursive meaning that it can be performed multiple times for obtaining better joint representations progressively. Extensive experiments on the public AVE dataset have shown that the proposed method achieves significantly better results than the state-of-the-art methods.
\end{abstract}
\section{Introduction}
Humans explore the surroundings with their advanced sensory system in daily life, e.g., eyes, ears, and noses. Heterogeneous information from various sensors floods into the human perceptual system, among which sound and vision are two dominant components. In multimodal machine learning, it turns out that the joint learning of audio and visual modalities usually achieves better performance than using single modality for various tasks, e.g., sound localization \cite{arandjelovic2018objects,hershey2000audio,owens2018audio,zhao2018sound,owens2016visually}, sound source separation \cite{ephrat2018looking,xu2019recursive,gao2018objectSounds,owens2018audio,zhao2018sound,khan2013speaker,yu2017permutation,gao2019co,rouditchenko2019self} and audio-visual event localization \cite{lin2019dual,tian2018audio,wu2019DAM}.

\begin{figure}[htbp]
    \centering
    \includegraphics[width=0.98\linewidth]
    {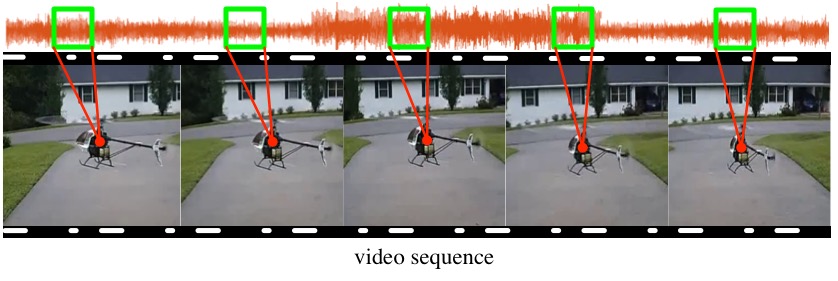}
    \caption{Audio-Visual Event (AVE) is an event both audible and visible. e.g., a person can see a helicopter in the visual sequence (the bottom row) and also hear the helicopter's engine sound in the audio sequence (the top row).}
    \vspace{-0.5cm}
    \label{fig:ave}
\end{figure}

In this paper, we focus on the audio-visual event localization task. 
As shown in Fig.~\ref{fig:ave}, an Audio-Visual Event (AVE) is defined in a video sequence that is both audible and visible. The audio-visual event localization task consists of two sub-tasks, one of which is to predict the event label while the other is to predict which segment of the video sequence has an audio-visual event of interest. As in the AVE definition, localizing an AVE must deal with heterogeneous information from both audio and visual modalities. 
Moreover, recent works~\cite{lin2019dual,tian2018audio,wu2019DAM} show that the performance after fusion outperforms the one that only uses a single modality. 
Although these approaches present interesting explorations, how to smartly fuse representations from both modalities is still a challenging task.

Multimodal fusion provides a global view of multiple representations for a specific phenomenon. To tackle the AVE localization problem, existing methods \cite{lin2019dual,tian2018audio} either fuse cell states out of LSTMs \cite{tian2018audio}, or fuse both hidden states and cell states from LSTMs \cite{lin2019dual}. Both aforementioned approaches exploit a plain multimodal fusion strategy, where the fusion results might be unstable as it is hard to guarantee good quality of the information used for the fusion, e.g., some noisy information from the background segments may also be included. 
Therefore, a more robust fusion strategy is needed for better representations. Wu \etal~\cite{wu2019DAM} introduce a cross-modal matching mechanism that exploits global temporal co-occurrences between two modalities and excludes the noisy background segments from the sequence. Intuitively, having global features to interact with local features would help to localize the event, but it needs additional supervision to manually filter the background segments. 

To summarize, existing methods either follow a straightforward multimodal fusion strategy (fuse both features directly), or require extra supervision (exclude background segments). 
Taking advantage of recursive fusion interactions between multimodal representations, we propose a novel joint co-attention fusion approach that is able to learn more robust representations with less supervision on excluding background segments.

Attention mechanism has been applied to many tasks~\cite{zhang2018self,tang2019multi,fu2019dual,xu2018structured,ding2020lanet,chen2020relevant,nam2017dual,tang2020xinggan}. For example, recent works in generative adversarial networks~\cite{zhang2018self,tang2019multi,tang2019attention} utilize a self-attention mechanism that relates different portions of a single image to compute a representation for itself. Besides self-attention, other works in Video Question Answering (VQA)~\cite{lu2016hierarchical,Nguyen_2018_CVPR} propose a co-attention mechanism, in which the image representation guides the text attention generation and in the meanwhile, the text representation also guides image attention generation. Moreover, both attention mechanisms allow attention-driven, long-range dependency modeling for their corresponding tasks. Motivated by these two attention techniques, we propose a new Joint Co-Attention (JCA) mechanism which develops on the basis of self-attention and co-attention. We utilize the joint representation to generate the attention masks for two uni-modalities while previous methods~\cite{lu2016hierarchical,Nguyen_2018_CVPR} independently generate attention mask for each other. In our approach, instead of using features from one single modality, each attention mask is generated using features from both modalities and thus it is more informative. As a result, each modality is attended not only by the features from itself (self-attended), but also by the features from the other modality (co-attended).

While the attention mechanism allows multimodal fusion in depth, we further introduce a double fusion mechanism, that can be integrated with attention mechanisms, allowing fusion both in depth and breadth. 
Existing works~\cite{lan2014multimedia,xu2015learning} exploit the double fusion to integrate representations from different modalities in a hybrid fusion manner, i.e., they fuse features using both early fusion (before feature embedding) and late fusion (after feature embedding). 
Different from existing methods~\cite{lin2019dual,tian2018audio,lan2014multimedia,xu2015learning} that fuse representations from multiple modalities simply by averaging, weighting or concatenation.
In this paper, we propose to integrate the double fusion method with our JCA mechanism.
First, the audio-guided attention~\cite{tian2018audio} is performed as early fusion. 
Then, we exploit Bi-LSTM~\cite{schuster1997bid} with residual embedding to extract features where we combine features before Bi-LSTM and after Bi-LSTM, leading to global temporal cues. After the Bi-LSTMs, the representations of two modalities are fused using the JCA mechanism as late fusion. 
Note that the JCA unit is recursive so that the joint co-attention process can be repeated for multiple times.

Overall, our contributions in this paper are summarized as follows:

\begin{itemize}
    \item We revisit the audio-visual event localization task and tackle the task from a multimodal fusion perspective which targets for better representations.
    \item We propose a novel joint co-attention mechanism and deploy it in deep audio-visual learning. It learns more robust representations by recursively performing fusions of the representations from two modalities.
    \item The integration of attention mechanism and double fusion method enables the model to learn long-range dependencies. 
    Extensive experiments show the superiority of our framework.
\end{itemize}
\section{Related Work}
\noindent\textbf{Audio-Visual Event Localization} aims to identify the event of interest in a video sequence and predict what category the event belongs to. 
Tian~\etal{}~\cite{tian2018audio} first define audio-visual event localization problem aiming to detect event which is both audible and visible. They design an audio-guided attention dual-LSTM network that captures each uni-modal representation and fuses them by concatenation for the final prediction. 
Lin \etal{}~\cite{lin2019dual} propose a dual-modality sequence-sequence framework that explores the global features of audio and visual modalities. Wu~\etal{}~\cite{wu2019DAM} introduce a dual attention matching mechanism that conducts cross-matching across modalities. They also leverage the global event feature by only considering segments containing audio-visual events, i.e., they filter out background segments to compute the global feature. However, determining background segments often requires more supervision. In our work, we propose to use less supervision to fulfill the task. Different from~\cite{tian2018audio}, we introduce a recursive layer that can be stacked and therefore recursively fuses two uni-modal representations multiple times to obtain more robust representations.

\noindent\textbf{Sound Localization} is to associate certain regions in a video that has the corresponding sound with visual-aid. To this end, Hershey \etal{}~\cite{hershey2000audio} use a Gaussian process model to measure the mutual information of the audio and visual motion. Owens and Efros~\cite{owens2018audio} propose a multi-sensory model that learns audio-visual correspondence in a self-supervised style to align the audio and visual frames and then localizes the sound source afterward. To investigate the correspondences between audio and visual components, Hu \etal{}~\cite{hu2019deep} propose a deep multimodal clustering network that adds similar parts among two modalities to the final output. Zhao \etal{}~\cite{zhao2018sound} propose PixelPlayer that learns to locate image regions by leveraging large amounts of unlabeled videos. Arandjelovic and Zisserman~\cite{arandjelovic2018objects} design two sub-networks that individually learn from audio tracks and image frames. After learning, they fuse two branches to predict correspondence. Senocak \etal{}~\cite{senocak2018learning} develop a localization module that is based on the attention mechanism to capture the correlation between audio and visual features. The attention mechanism they adopt is quite plain where they transpose visual embedding and then multiply with audio embedding.

\noindent\textbf{Multimodal Attention} involves interaction at least two features from different modalities. In Video Question Answering (VQA), Lu~\etal{}~\cite{lu2016hierarchical} propose a hierarchical attention technique that co-attends to the features extracted from text language modality and visual modality. Another work in VQA, Nguyen and Okatani~\cite{Nguyen_2018_CVPR} introduce a memory-based co-attention technique that enables dense interactions between the two modalities, and then both modalities contribute to the selection of the right answer. In emotion recognition, Zadeh \etal{}~\cite{zadeh2018memory} exploit a small neural network that takes the concatenated cell states of three different LSTMs for language, audio and visual components as input and then output the attended features. Wang \etal{}~\cite{wang2019words} present an attention gating mechanism where they try to learn a nonverbal shift vector by weighting features from different modalities. Different from the two aforementioned work that only perform attention operation once, we develop a joint co-attention mechanism that can be recursively performed.

\noindent\textbf{Multimodal Fusion} also known as the integration of information from multiple modalities~\cite{baltruvsaitis2018multimodal}, allows for more robust representations by leveraging multiple modalities and it can be categorized into two types: model-agnostic and model-based. Here, we only review the model-agnostic approaches, i.e., early fusion~\cite{snoek2005early}, late fusion~\cite{ramirez2011modeling,snoek2005early} and hybrid fusion~\cite{lan2014multimedia,xu2015learning} as it is more related to our work. Early fusion combines low-level features of each modality while late fusion uses uni-modal decision values based on a fusion algorithm, e.g., averaging, weighting. Hybrid fusion, or double fusion, attempts to take advantage of both early and late fusion mechanisms. It has been widely used in the research of multimodal learning, e.g., multimodal speech recognition~\cite{wu2006multi,sun2016look} and multimedia event detection~\cite{lan2014multimedia,xu2015learning,jiang2012leveraging}. Besides a bigger picture of fusion strategies, there are many specific fusion strategies in terms of feature-level. Rahman \etal{}~\cite{rahman2019watch} adopt element-wise addition or multiplication, channel-wise concatenation, and fully-connected neural network to fuse information from three different modalities: language, audio, and visual modality.

\section{Approach}
\begin{table}[!t] \small
    \centering
    \caption{Main symbols used throughout the paper.}
    \begin{tabular}{p{1.2cm}p{6.2cm}}
        \toprule
        \textbf{Symbol} & \textbf{Definition} \\
        \hline
        $\mathcal{S}_a/\mathcal{S}_v$ & audio/visual sequence\\
        $s^t_a/s^t_v$ & audio/visual $t$-th segment-level feature\\
        $f^t_a/f^t_v$ & audio/visual feature after re-representation layer\\
        $\mathbf{A}/\mathbf{V}/\mathbf{J}$ & audio/visual/joint sequence-level feature\\
        $\mathbf{C}_a/\mathbf{C}_v$ & joint-audio/joint-visual affinity matrix\\
        $d_a/d_v/d$ & dimmesion of audio/visual/joint feature\\
        $\ell$ & recursive times of joint co-attention layer\\
        $\mathbf{H}_a/\mathbf{H}_v$ & audio/visual feature after joint co-attention layer\\
        $\mathbf{W}_{ja}/\mathbf{W}_{jv}$ & parameters between ($\mathbf{J}$ and $\mathbf{A}$)/($\mathbf{J}$ and $\mathbf{V}$)\\
        $\mathbf{W}_{a}/\mathbf{W}_{v}$ & parameters for feature $\mathbf{A}/\mathbf{V}$\\
        $\mathbf{W}_{ca}/\mathbf{W}_{cv}$ & parameters for feature $\mathbf{C}_a/\mathbf{C}_v$\\
        $\mathbf{W}_{h_a}/\mathbf{W}_{h_v}$ & parameters for feature $\mathbf{H}_a/\mathbf{H}_v$\\
        \bottomrule
    \end{tabular}
    \vspace{-15pt}
    \label{tab:notation}
\end{table}
In this section, we introduce the overall architecture of our proposed joint co-attention network for the supervised audio-visual event localization task layer by layer, as shown in Fig.~\ref{fig:arch}. To start with the description, we first set forth the notations in Sec.~\ref{sec:pre}, then the sequence feature re-representation layer is described in Sec.~\ref{sec:Re-Rep}. Next, we introduce the proposed joint co-attention layer in Sec.~\ref{sec:co-attn}. Lastly, we explain the final prediction layer in Sec.~\ref{sec:prediction}.

\subsection{Notations}
\label{sec:pre}
The symbols used throughout the paper are listed in Table~\ref{tab:notation}. An Audio-Visual Event (AVE) is defined as an event that is both visible and audible~\cite{tian2018audio}. As in \cite{tian2018audio,wu2019DAM}, for a given audio-visual video sequence $\mathcal{S}\!=\!(\mathcal{S}_a$,~$\mathcal{S}_v\!)$, while $\mathcal{S}_a$ denotes the audio portion and $\mathcal{S}_v$\,denotes the visual portion. The video sequence $\mathcal{S}$ is split into $N$ non-overlapping yet continuous segments where each segment is typically one second long. For each segment, a label $y \in \left\{0, 1\right\}$ is given, while 0 indicates the segment is background and 1 indicates that is an AVE. The sequence features, i.e., $\mathcal{S}_a$ and $\mathcal{S}_v$ are extracted using a pre-trained CNN. We denote the extracted segment-level feature as $s^t_a$ and $s^t_v$ corresponding to the audio and visual modality respectively, where $t \in \{1, 2, \cdots, N\}$. Our network is built on the basis of fixed $s^t_a$ and $s^t_v$.

\subsection{Re-Representation Layer}
\label{sec:Re-Rep}
\begin{figure*}[!t] \small
    \centering
    \includegraphics[width=0.9\linewidth]
    {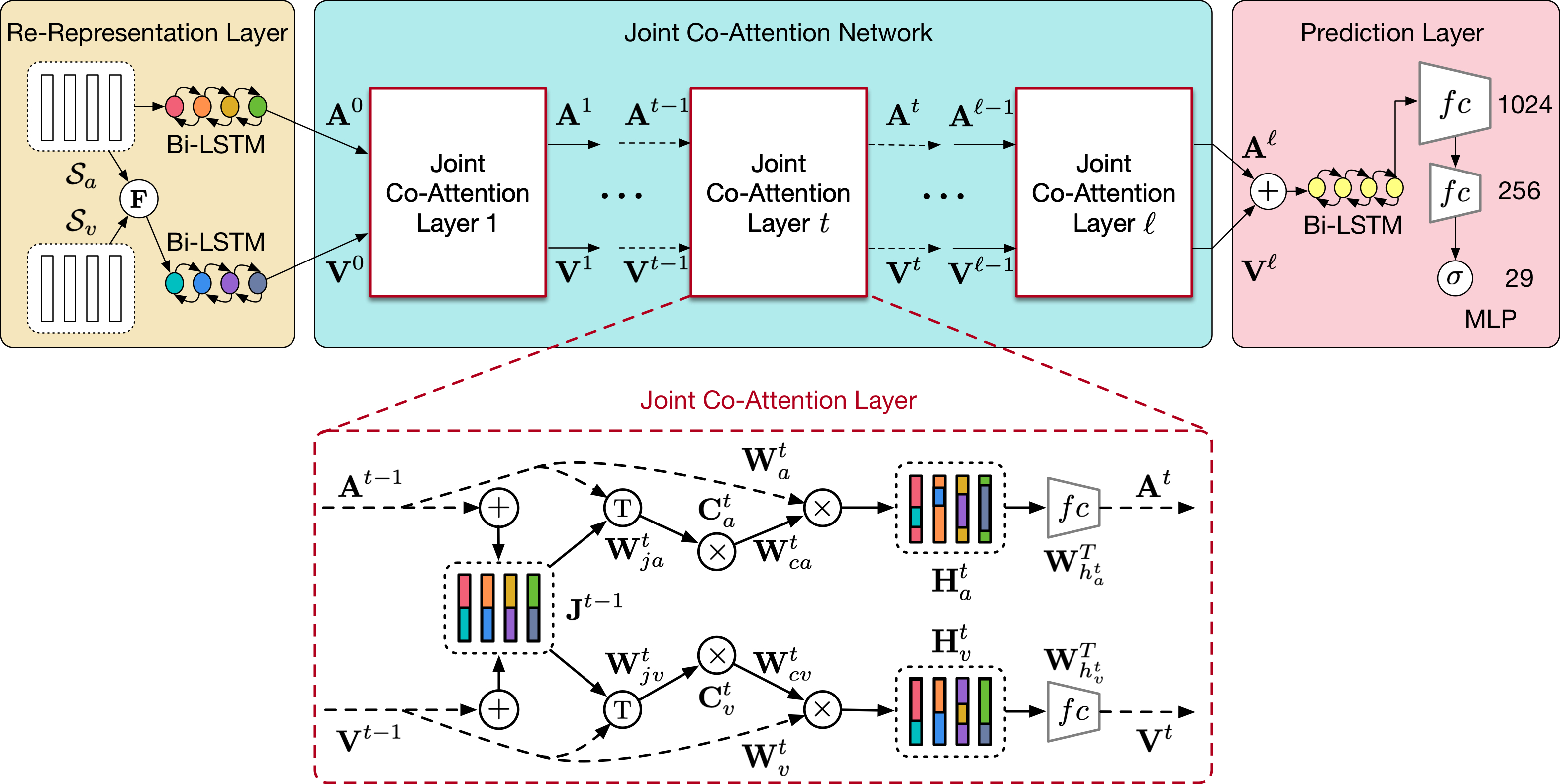}
    \caption{The overall structure of the proposed framework. We split it into three parts, i.e., sequence feature re-representation layer, joint co-attention network and category prediction layer. For the symbols, \textcircled{+} denotes concatenation, \textcircled{\tiny{F}} denotes early fusion of audio feature and visual feature, \textcircled{$\sigma$} denotes the softmax function, \textcircled{\tiny{T}} is transpose operator, and \textcircled{$\times$} is matrix multiplication operator.}
    \label{fig:arch}
\end{figure*}

Sequence representation contains temporal cues among the sequential stream, and LSTM has shown its superiority in learning those temporal cues. Therefore, we use the LSTM to modulate the sequence representations. Different from existing methods~\cite{lin2019dual,tian2018audio}, we add a residual embedding to the output of the LSTM in order to produce better representation.
The structure of the proposed re-representation layer is shown in Fig.~\ref{fig:arch}.

\noindent\textbf{Audio Representation.} In general, a sequence of audio feature $\mathbf{A}$ contains $N$ continuous segments, i.e., \{$s^1_a, s^2_a, \cdots, s^N_a$\}, where each $s^t_a$ is a $128\times1$ dimensional vector. We adopt Bi-directional LSTM~\cite{schuster1997bid} with residual embedding, in our case, concatenation, to learn the audio representation:
\begin{equation}
    \stackrel\longrightarrow{f^t_a} = \text{Bi-LSTM}(\left[
        \begin{array}{c}
           \stackrel\longrightarrow{f^{t-1}_a}\\
           s^t_a
        \end{array}
        \right]),
    \label{eqa:1}
\end{equation}
\begin{equation}
    \stackrel\longleftarrow{f^t_a} = \text{Bi-LSTM}(\left[
        \begin{array}{c}
           \stackrel\longleftarrow{f^{t+1}_a}\\
           s^t_a
        \end{array}
        \right]),
    \label{eqa:2}
\end{equation}
where the arrow indicates the direction of information flowing. Therefore, $f^t_a =$ \text{concat}$(\stackrel\longleftarrow{f^t_a},\stackrel\longrightarrow{f^t_a})$. We concatenate $N$ segments along the time axis and create a new feature matrix, i.e., $\mathbf{A} = \text{concat}(f^1_a,\ldots, f^N_a) \in \mathbb{R}^{N\times{}d_a}$.

\noindent\textbf{Visual Representation.} Unlike audio features which are 1D features, visual features are 2D features extracted from image frames. This brings problems as the model needs to process two types of features with different dimensions simultaneously. Typically, the size of each visual feature is $512\times7\times7$ as in \cite{tian2018audio,wu2019DAM}. If we simply conduct pooling in the height and width dimensions and reduce them into size 1 ($7\!\rightarrow\!1$), the performance of this reduction procedure can barely be guaranteed as the stride is big and may leave out useful information. Further study about the influence of applying different pooling methods is conducted in the ablation studies in Sec.~\ref{sec:ablation}. 
To smoothly reduce the dimension of raw visual features, we obtain the scaled dot-product of audio feature and visual feature for each segment. We denote the scaled dot-product as $f^t_v$.
Then we follow a similar routine to encode the sequence of visual features like the audio sequence using LSTM with residual embedding. Consequently, we use a matrix for visual representation $\mathbf{V} = \text{concat}(f^1_v,\ldots, f^N_v) \in \mathbb{R}^{N\times{}d_v}$.

\subsection{Joint Co-Attention Network}
\label{sec:co-attn}
We now introduce the Joint Co-Attention (JCA) layer as shown in Fig.~\ref{fig:arch}. The proposed joint co-attention layer attends to visual features and audio features simultaneously. It takes the audio representation $\mathbf{A}$ and the visual representation $\mathbf{V}$ as inputs and concatenates two representations as the joint representation $\mathbf{J}$. We employ $\mathbf{J}$ to co-attend to $\mathbf{A}$ and $\mathbf{V}$, respectively. It is worth noting that we only preserve $\mathbf{J}\!\rightarrow\!{\mathbf{A}}$ (i.e., joint feature attend to audio feature) and $\mathbf{J}\!\rightarrow\!{\mathbf{V}}$ (i.e., joint feature attend to visual feature), the inverse directions of $\mathbf{A}\!\rightarrow\!{\mathbf{J}}$ and $\mathbf{V}\!\rightarrow\!{\mathbf{J}}$ are abandoned for simplicity, which is different from the original co-attention mechanism~\cite{lu2016hierarchical}. One property of JCA is mutual attention, that is, it can attend to features from two different modalities simultaneously. Another special property of JCA is stackability, i.e., we can stack several JCAs so that we can recursively perform the process multiple times. Extensive experiments on different recursive times of the JCA unit are shown in Sec.~\ref{sec:ablation}.

\noindent\textbf{Primary Idea for Joint Co-Attention.} Recent studies~\cite{lu2016hierarchical,Nguyen_2018_CVPR} explore the co-attention theory in Visual Question Answering (VQA). The text sequence representations and the visual sequence representations attend mutually to obtain new representations. Inspired by this, we explore a mode that allows representation from one modality not only attending to the other representation from the other modality but also attending to the representation from its original modality. Given audio representation $\mathbf{A} \in \mathbb{R}^{N\times d_a}$, and visual representation $\mathbf{V} \in \mathbb{R}^{N\times d_v}$, the joint representation $\mathbf{J} \in \mathbb{R}^{N\times d}$ is acquired by the concatenation of $\mathbf{A}$ and $\mathbf{V}$, i.e, $    \mathbf{J} = \left[
                \begin{array}{c}
                    \mathbf{A}\\
                    \mathbf{V}
                \end{array}
                \right]$, 
where $d = d_a+d_v$. We take audio feature $\mathbf{A^\ell}$ as an example to elaborate the process of joint co-attention. Here, we denote $\mathbf{A}^1$ as the initial state of audio feature and $\mathbf{A}^\ell$ as the audio feature after $\ell$-th joint co-attention layer. First, the $(\ell\!-\!1)$-th layer's audio representation $\mathbf{A}^{\ell-1}$ is concatenated with $\mathbf{V}^{\ell-1}$ to obtain joint representation $\mathbf{J}^{\ell-1}$; next, we employ the $\mathbf{J}^{\ell-1}$ to attend to $\mathbf{A}^{\ell-1}$ and finally obtain the $\ell$-th layer's audio feature $\mathbf{A}^{\ell}$. Following the similar rules, the new visual feature $\mathbf{V}^{\ell}$ is obtained.

\noindent\textbf{Learning to Joint Co-Attend.} Fusion is one of the key challenges for multimodal learning~\cite{baltruvsaitis2018multimodal}. Following recent studies~\cite{lu2016hierarchical,Nguyen_2018_CVPR} in VQA, we specifically derive the fusion to fit our audio-visual event localization task. After calculating the joint representation matrix $\mathbf{J}$, we use it to attend to different uni-modal representations via the following equation:
\begin{equation}
    \mathbf{C}_a = \text{Tanh}\left(\frac{\mathbf{A}^\mathrm{T}\mathbf{W}_{ja}\mathbf{J}}{\sqrt{d}}\right),
    \label{eqa:4}
\end{equation}
where $\mathbf{C}_a$ is the joint-audio affinity matrix, $\mathrm{T}$ denotes transpose operation, and $\mathbf{W}_{ja}\in \mathbb{R}^{N\times N}$ is a learnable weight matrix ($\mathbf{W}_{ja}$ is implemented as fully-connected layer). Following the same rule, the joint-visual affinity matrix $\mathbf{C}_v$ can be written as
\begin{equation}
    \mathbf{C}_v = \text{Tanh}\left(\frac{\mathbf{V}^\mathrm{T}\mathbf{W}_{jv}\mathbf{J}}{\sqrt{d}}\right),
    \label{eqa:5}
\end{equation}
where $ \mathbf{W}_{jv} \in \mathbb{R}^{N\times N}$ is also a learnable weight matrix. After calculating the joint uni-modal affinity matrices $\mathbf{C}_a$ and $\mathbf{C}_v$, we then calculate the attention probabilities map $\mathbf{H}_a, \mathbf{H}_v$ of two modalities as, $\mathbf{H}_a = \text{ReLU}\left(\mathbf{W}_{a}\mathbf{A} + \mathbf{W}_{ca}\mathbf{C}_a^\mathrm{T} \right)$ and $\mathbf{H}_v = \text{ReLU}\left(\mathbf{W}_{v}\mathbf{V} + \mathbf{W}_{cv}\mathbf{C}_v^\mathrm{T} \right)$, 
where $\mathbf{H}_a \in \mathbb{R}^{k\times d_a}, \mathbf{H}_v \in \mathbb{R}^{k\times d_v}$ represent the attention probabilities map of audio modality and visual modality, respectively. $\mathbf{W}_a, \mathbf{W}_v \in \mathbb{R}^{k \times N}$, $\mathbf{W}_{ca}, \mathbf{W}_{cv} \in \mathbb{R}^{k\times d}$ are learnable weight matrices.

After obtaining the attention map $\mathbf{H}_a$ and $\mathbf{H}_v$, we recompute the new audio representation and new visual representation by
\begin{equation}
    \mathbf{A}^\ell = g(\mathbf{A}^{\ell-1}, \mathbf{W}^\mathrm{T}_{h^{\ell}_a}\mathbf{H}^{\ell}_a),
    \label{eqa:8}
\end{equation}
\begin{equation}
    \mathbf{V}^\ell = g(\mathbf{V}^{\ell-1}, \mathbf{W}^\mathrm{T}_{h^{\ell}_v}\mathbf{H}^{\ell}_v),
    \label{eqa:9}
\end{equation}
where $\mathbf{W}_{h^{\ell}_a}, \mathbf{W}_{h^{\ell}_v} \in \mathbb{R}^{k\times N}$ are learnable weight matrices in the $\ell$-th layer. $\ell\!-\!1$ represents the features produced by the $\ell\!-\!1$-th layer. In our case, $g$ is a summation function.

\noindent\textbf{Fusion by Fusion.} Multimodal fusion can generate more robust representation using the features from multiple modalities that are collected for the same phenomenon. Earlier studies~\cite{lin2019dual,tian2018audio,wu2019DAM} particularly exploit the method in an audio-visual dual-modality setting either directly fusing the features or using cross dot product operation. Different from them, we consider multimodal fusion as a recursive process, where we fuse audio representation $\mathbf{A}$ and visual representation $\mathbf{V}$ recursively to obtain more robust representations. Following Eq.~\eqref{eqa:8} and Eq.~\eqref{eqa:9}, we generalize this recursive process as
\begin{equation}
    \mathbf{A}^\ell = g(\cdots g(\mathbf{A}^0, \mathbf{W}^\mathrm{T}_{h^1_a}\mathbf{H}^1_a) \cdots, \mathbf{W}^\mathrm{T}_{h^{\ell}_a}\mathbf{H}^{\ell}_a),
    \label{eqa:10}
\end{equation}
\begin{equation}
    \mathbf{V}^\ell = g(\cdots g(\mathbf{V}^0, \mathbf{W}^\mathrm{T}_{h^1_v}\mathbf{H}^1_v) \cdots, \mathbf{W}^\mathrm{T}_{h^{\ell}_v}\mathbf{H}^{\ell}_v),
    \label{eqa:11}
\end{equation}
where $\ell$ represents the amount of times that the joint co-attention is repeated. After fusing $\ell$ times, we will obtain two more robust representations for audio and visual modality, respectively.

\subsection{Prediction Layer}
\label{sec:prediction}
The audio-visual event localization task aims to identify an AVE in a given video sequence and predict which category the AVE belongs to. Note that the input sequences of different categories and the backgrounds are heterogeneous. As a consequence, it is even harder to complete the task. Different from~\cite{wu2019DAM}, we use less supervision by only considering event category labels. Before prediction, early fusion of two separate modalities is performed, and then two uni-modal representations are re-represented as $\mathbf{A}$ and $\mathbf{V}$. Next, following the fusion method in Sec.~\ref{sec:co-attn} of joint co-attention, we fuse two uni-modal representations multiple times to get $\mathbf{A}^\ell$ and $\mathbf{V}^\ell$. Finally, $\mathbf{A}^\ell$ and $\mathbf{V}^\ell$ are taken as input into the final category prediction layer:
\begin{equation}
    \text{prediction} = \text{MLP}\bigg(\text{Bi-LSTM}\Big(\left[\!
                \begin{array}{c}
                    \mathbf{A}^\ell\\
                    \mathbf{V}^\ell
                \end{array}\!
                \right]\Big)\bigg).
\end{equation}
where MLP denotes Multilayer Perceptron and Bi-LSTM is to modulate audio and visual representations jointly. In experiments, the MLP is implemented by using a two-layer fully-connected network embedded with 1,024/256 hidden units and a Sigmoid layer $\sigma$, as shown in Fig.~\ref{fig:arch}. After that, the predicted category is the one that corresponds to the max value in the prediction vector. During training, we use the Multi Label Soft Margin loss function to optimize the entire network.

\section{Experiments}
\subsection{Experimental Setup}
\label{sec:exp}
\begin{figure*}[!t] \small
    \centering
    \includegraphics[width=0.9\linewidth]
    {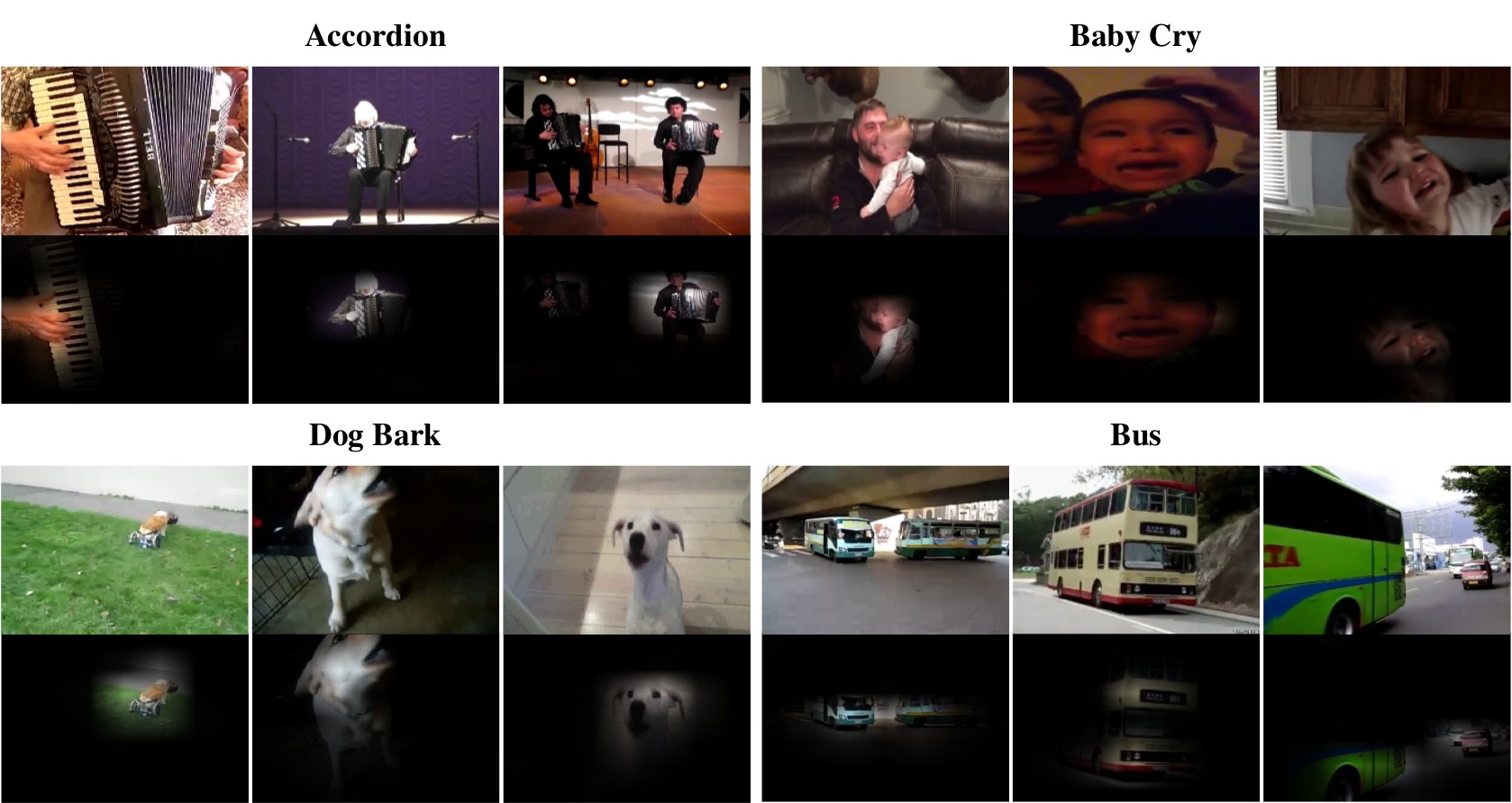}
    \caption{Visualizing attention mask of the proposed joint co-attention mechanism on four categories of the AVE dataset.} 
    \vspace{-0.4cm}
    \label{fig:results2}
\end{figure*}

\noindent\textbf{Audio-Visual Event Dataset.} The Audio-Visual Event (AVE) dataset by \cite{tian2018audio} is a subset of AudioSet~\cite{Jort2017audioset}. It consists of $4,143$ video clips that involve $28$ event categories. We adopt the split technology of \cite{tian2018audio} where train/validation/test sets are $3,309/402/402$ video clips, respectively. 
While training, the model has no access to the test portion to better evaluate the model’s generalization ability. 
For the AVE dataset, it contains comprehensive audio-visual event types, in general, instrument performances, human daily activities, vehicle activities, and animal actions. To be more specific, for more detailed event categories, take instrument performances as an example, AVE dataset contains accordion playing, guitar playing, and ukulele playing, etc. A typical video clip is 10 seconds long and is labeled with the start point and endpoint at the segment level to clarify whether the segment is an audio-visual event. Sample images and their attended images are shown in Fig.~\ref{fig:results2}.

\noindent\textbf{Evaluation Metrics.} 
We follow \cite{lin2019dual,tian2018audio,wu2019DAM} and adopt the global classification accuracy obtained from the last prediction layer  as the evaluation metric.
For an input video sequence, our goal is to predict the category label for each segment. It is worth noting that the background category contains 28 backgrounds since each event category can have its own background so that it is hard to predict. 

\noindent\textbf{Experimental Details.} Following~\cite{tian2018audio,wu2019DAM}, we adopt pre-trained CNN models to extract features for each audio and visual segment. Specifically, we exploit the VGG19~\cite{simonyan2014very} network pre-trained on ImageNet~\cite{deng2009imagenet} as the backbone to extract segment-level visual features. Meanwhile, for the audio segment, we extract the segment feature using a Vggish network~\cite{hershey2017cnn} which is pre-trained on AudioSet~\cite{Jort2017audioset}. For a fair comparison, we use the same extracted features (i.e, audio and visual features) as used in~\cite{tian2018audio,wu2019DAM}. In the training stage, the only supervision we exploit is the annotation labels for the temporal segments.
\subsection{Comparison with Existing Methods}
\label{sec:comparison}

\noindent\textbf{State-of-the-Art Comparison.} Results compared with the leading methods are reported in Table~\ref{tab:comparison}. We take a similar model architecture as in \cite{tian2018audio} and run single modality models as our baselines, which only take audio features or visual features during the experiments. First, to validate the proposed method can enable efficient interactions between audio features and visual features, we compare with a state-of-the-art temporal labeling network, i.e, ED-TCN~\cite{lea2017temporal}, which can integrate information from multiple temporal segments. Next, to verify the effectiveness of our fusion strategy of audio feature and visual feature, we compare with two methods, i.e,  Audio-Visual~\cite{tian2018audio} and AVSDN~\cite{lin2019dual}. Both methods utilize a straightforward fusion strategy, where fuses the audio and visual features out of LSTMs by concatenation. Lastly, to evaluate that our method is tolerant with less supervision, we compare our method with DAM~\cite{wu2019DAM},  which needs additional supervision to exclude event-irrelevant segments during training.

\noindent\textbf{Comparison Analysis.} Due to the absence of interactions between audio modality and visual modality, our proposed model can easily surpass the performance of the baselines. In addition, by comparing with ED-TCN, our model enables more effective interactions between two modalities. Thus, it can be testified that interactions or fusion can boost the task performance and our model is more superior on enabling interactions between two different modalities. Unsurprisingly, by fusing the two different features using our joint co-attention mechanism, our model outperforms Audio-Visual and AVSDN using a plain fusion strategy. Moreover, even without additional effort to exclude event-irrelevant segments, our model can learn useful representations from noisy inputs and contribute to better performance.
\begin{table}[!t]
    \centering
    \caption{Results of comparisons with the state-of-the-art methods on the AVE dataset. For a fair comparison, * is obtained by exploiting the same pre-trained audio and visual features. While the task is hard, it can still be observed that our model outperforms the existing methods.}
    \begin{tabular}{p{4.5cm}c}
        \toprule
        \textbf{Method}  &   \textbf{Accuracy (\%)}\\
        \hline
        Audio\_Only (Vggish~\cite{hershey2017cnn})&   59.5 \\
        Visual\_Only (Vgg19~\cite{simonyan2014very})&  55.3 \\
        ED-TCN~\cite{lea2017temporal}&   46.9\\
        Audio-Visual~\cite{tian2018audio}& 71.4 \\
        AVSDN*~\cite{lin2019dual}& 72.6 \\
        Full-Audio-Visual~\cite{tian2018audio}& 72.7\\
        DAM~\cite{wu2019DAM}&   74.5\\
        \textbf{Ours} & \textbf{76.2}\\
        \bottomrule
    \end{tabular}
     \vspace{-10pt}
    \label{tab:comparison}
\end{table}
\subsection{Ablation Study}
\label{sec:ablation}
\begin{table}[!t] \small
    \centering
    \caption{Ablation studies on the proposed framework. Uni-modal Bi-LSTM is the LSTM in sequence feature re-representation layer while Joint Bi-LSTM is the one in the prediction layer. * denotes we remove the residual embedding of LSTMs while \dag~denotes that we adopt the primary co-attention mechanism into the proposed framework.}
    \begin{tabular}{p{4.5cm}c}
        \toprule
        \textbf{Model}  &   \textbf{Accuracy (\%)}\\
        \hline
        Ours w/o Uni-modal Bi-LSTM    &  74.5 \\
        Ours w/o Joint Bi-LSTM    &  74.9 \\
        Ours w/o Residual Embedding*    &  75.2 \\
        Ours w/ GRU~\cite{cho2014learning} &   75.3\\
        Ours w/ Average Pooling   &   75.1\\
        Ours w/ Max Pooling   &   75.0\\
        Ours w/ Co-Attention\dag~\cite{Nguyen_2018_CVPR}    &   75.4\\
        Ours w/ Joint Co-Attention  &  \textbf{76.2} \\
        \bottomrule
    \end{tabular}
    \vspace{-10pt}
    \label{tab:ablation}
\end{table}

\begin{table}[!t] \small
    \centering
    \caption{Results of different fusion strategies to generate joint representation ${\mathbf{J}}$.}
    \begin{tabular}{p{2.6cm}cc}
        \toprule
        \textbf{Strategy} & \textbf{Accuracy (\%)}&\textbf{Params} \bm{$\times10^6$}   \\
        \hline
        Addition    &  75.0 & \textbf{22.67}\\
        Multiplication   &  74.6 & \textbf{22.67}\\
        Concatenation    &  75.2 & 22.72\\
        Addition + FC &   75.5 & 22.78\\
        Multiplication + FC &   75.3 & 22.78\\
        Concatenation + FC &   \textbf{76.2} & 22.83\\
        \bottomrule
    \end{tabular}
    \vspace{-15pt}
    \label{tab:fusion}
\end{table}

\begin{table}[!t] \small
    \centering
    \caption{Variations on the proposed JCA architecture. Unlisted value are identical to those of the first row of the model. Besides accuracy, we also calculate the parameters of each setting.}  
    \begin{tabular}{p{1.9cm}p{0.1cm}p{0.3cm}p{0.1cm}cc}
        \toprule
        &\bm{$d_a$}&\bm{$d_v$} &\bm{$\ell$}&\textbf{Accuracy (\%)} & \textbf{Params} \bm{$\times10^6$}\\
        \hline
        (A)~$256\times1024$&256&1024&4&{75.4}&{50.7}\\
        \hline
        \multirow{5}{*}{(B)~$512\times512$}&512&512&1&{75.1}&{14.9}\\
        &&&2&{75.4}&{17.5} \\
        &&&3&{75.6}&{20.1} \\
        &&&4&{\textbf{76.2}}&{22.8}\\
        &&&5&{75.6}&{25.4}\\
        \hline
        \multirow{4}{*}{(C)~$256\times256$}&256&256&2&{74.8}&{4.6}\\
        &&&3&{75.0}&{5.2} \\
        &&&4&{75.1}&{5.9} \\
        &&&5&{74.8}&{6.6}\\ 
        \hline
        (D)~$128\times128$&128&128&4&{73.8}&\textbf{1.6}\\ 
        \bottomrule
    \end{tabular}
    \vspace{-10pt}
    \label{tab:lnums}
\end{table}

\noindent\textbf{Framework Decoupling.} We break down the proposed framework and evaluate them separately in different settings, as shown in Table \ref{tab:ablation}. For the Bi-LSTM, we define it as two types, one in sequence feature re-representation as uni-modal Bi-LSTM while the other in the prediction layer as joint Bi-LSTM. Experiments on two Bi-LSTMs are denoted as `Ours w/o Uni-modal Bi-LSTM' and `Ours w/o Joint Bi-LSTM, respectively. 

Moreover, GRU~\cite{cho2014learning} is used as an alternative to Bi-LSTM for further investigation, denoted as `Ours w/ GRU'. For early fusion, we evaluate two direct pooling methods i.e., global average pooling and global max pooling, denoted as `Ours w/ Average Pooling' and `Ours w/ Max Pooling', respectively. Lastly, the `Ours w/ Co-Attention' represents that we replace joint co-attention with the original co-attention~\cite{lu2016hierarchical}. Our full model is denoted as `Ours w/ Joint Co-Attention'.

\noindent\textbf{Framework Analysis.} Results are showed in Table~\ref{tab:ablation}. First, the overall performance of the proposed framework outperforms the state-of-the-art method~\cite{wu2019DAM} which needs additional supervision. Among all the observed declines, Bi-LSTM has the highest impact. That confirms the effectiveness of the Bi-LSTM part. For alternatives to early fusion, neither the global average pooling nor the global max pooling surpasses our full model. 

Among the experiment results with two different co-attention mechanisms, i.e., original co-attention method~\cite{Nguyen_2018_CVPR} and our joint co-attention method, our joint co-attention method excels the original co-attention method which follows a dual-modality mutual attending way (visual features attend to audio features and audio features attend to visual features). By not only attending to the corresponding modality but also the modality of itself, our proposed joint co-attention method performs better in the audio-visual fusion task. To sum up, the ablation studies demonstrate the efficiency of our proposed framework.

\noindent\textbf{Studies on Different Fusion Strategies.} To further investigate how different fusion strategies used to produce joint representation ${\mathbf{J}}$ can influence the performance of the proposed model, we exploit various fusion strategies as variations of our proposed model. (i) Element-wise addition; (ii) Element-wise multiplication; (iii) Channel-wise concatenation; (iv) Fully-connected neural network (FC). 

Results are showed in Table~\ref{tab:fusion}. It can be witnessed that directly making concatenation, addition or multiplication impair the performance of the fusion representation while introducing FC can slightly increase numbers of the parameters (less than 0.7\%), but the performance can increase by 2.1\%.

\begin{figure*}[!t] \small
    \centering
    \includegraphics[width=0.9\linewidth]
    {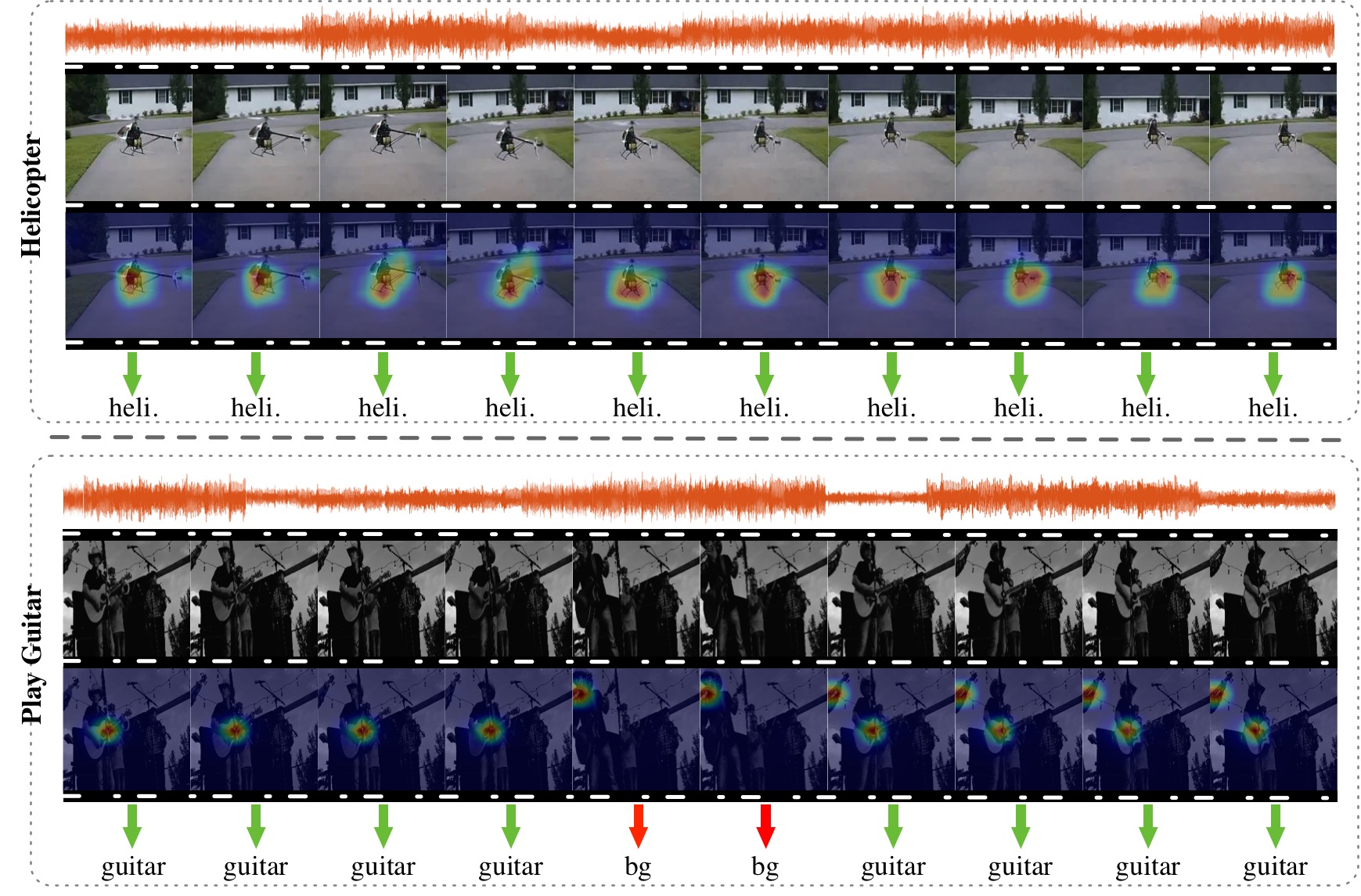}
    \caption{Two qualitative results on audio-visual localization task. The first example is helicopter hovering, i.e, `heli.' is the abbreviation for helicopter for better layout; while second example is playing guitar, i.e., `guitar' for short, `bg' denotes `background'. The green arrow represents the correct prediction whereas the red arrow denotes the wrong prediction. To visualize where they attend to, we generate images with their corresponding attention map. \textit{Best viewed in color}.}
    \label{fig:results}
    \vspace{-0.4cm}
\end{figure*}

\noindent\textbf{Studies on Recursive Times $\ell$, and Dimensions of $d_a$ and $d_v$.} To further investigate the proposed framework, we vary the proposed model in different ways and then evaluate the accuracy under each circumstance. The results are presented in Table~\ref{tab:lnums}. 

The $\ell$ denotes the times of JCA that are recursively performed whereas $d_a$ and $d_v$ denote the dimension of audio feature and visual feature, respectively. We observe that reducing the dimensions of the input features ($d_a$ and $d_v$) hurts the model's performance, which suggests high-dimensional features may be suitable for the fusion. However, features with extremely high dimension would bring a lot of computation. In practice, one should make a trade-off here. If we only look at row (B) or row (C), it is easy to find that as the recursive times of JCA increase, the performance improves. This also validates our motivation that repeating the fusion process helps our model to learn more robust representations.


\subsection{Qualitative Evaluation}
\label{sec:evalution}
In this section, we show some qualitative results of our proposed framework in Fig.~\ref{fig:results}. For each row in Fig.~\ref{fig:results}, the left is the category of this audio-visual event; the top content is the waveform of input audio sequence; the middles are raw frames and frames with attention map of the input video sequence; the bottom is the audio-visual event prediction. 

Among the two instances in Fig.~\ref{fig:results}, the second instance is much harder as the scene is more complicated where different people are playing different instruments. In the beginning, the proposed network predicts well. However, as the singer changing his posture, the guitar can hardly be seen even with our eyes. Therefore, the network fails to predict it as playing guitar. Surprisingly, as the singer turns back to the front, our network works again, and it marks two guitars in the picture even the other guitar is indistinct. 
More results are shown in Fig. \ref{fig:results2}.
We can see that the proposed co-attention model adaptively captures different sound sources in different semantic regions, such as accordion, crying boy/girl/babies, barking dog, horning bus, ukulele, etc.

\section{Conclusion}
In this paper, we investigate an interesting problem on deep audio-visual learning for the AVE task. To better cope with this multimodal learning task, we propose a novel joint co-attention mechanism with double fusion. To the best of our knowledge, this is the first time of applying the co-attention mechanism into the audio-visual event localization task. The integration with double fusion leading to better representations for the AVE task by co-attending to both audio and visual modalities. Moreover, experimental results on the AVE dataset have confirmed the superiority of the proposed framework.

{\small
\bibliographystyle{ieee_fullname}
\bibliography{egbib}
}

\end{document}